\documentclass{article}

\usepackage{hyperref}
\usepackage{amssymb}
\usepackage{amsmath}
\usepackage{url}
\usepackage{graphicx}
\usepackage{subcaption}
\usepackage{caption}
\usepackage{booktabs}
\usepackage{multirow}
\usepackage{microtype}
\usepackage{amsfonts}
\usepackage{booktabs} 

\usepackage{hyperref}

\usepackage[accepted]{icml2018}

\icmltitlerunning{CoVeR: Learning Covariate-Specific Vector Representations with Tensor Decompositions}

\begin{document}

\twocolumn[
\icmltitle{CoVeR: Learning Covariate-Specific Vector Representations with Tensor Decompositions}



\icmlsetsymbol{equal}{*}

\begin{icmlauthorlist}
\icmlauthor{Kevin Tian}{equal,CS}
\icmlauthor{Teng Zhang}{equal,MSE}
\icmlauthor{James Zou}{BMDS}
\end{icmlauthorlist}

\icmlaffiliation{CS}{Department of Computer Science,
Stanford University}
\icmlaffiliation{MSE}{Department of Management Science and Engineering, Stanford University}
\icmlaffiliation{BMDS}{Department of Biomedical Data Science
Stanford University}

\icmlcorrespondingauthor{Kevin Tian}{kjtian@stanford.edu}

\icmlkeywords{Word embedding, tensor decomposition, covariate}

\vskip 0.3in
]



\printAffiliationsAndNotice{\icmlEqualContribution} 

\begin{abstract}
Word embedding is a useful approach to capture co-occurrence structures in large text corpora. However, in addition to the text data itself, we often have additional covariates associated with individual corpus documents---e.g. the demographic of the author, time and venue of publication---and we would like the embedding to naturally capture this information. We propose CoVeR, a new tensor decomposition model for vector embeddings with covariates. CoVeR jointly learns a \emph{base} embedding for all the words as well as a weighted diagonal matrix to model how each covariate affects the base embedding. To obtain author or venue-specific embedding, for example, we can then simply multiply the base embedding by the associated transformation matrix. The main advantages of our approach are data efficiency and interpretability of the covariate transformation. Our experiments demonstrate that our joint model learns substantially better covariate-specific embeddings compared to the standard approach of learning a separate embedding for each covariate using only the relevant subset of data, as well as other related methods. Furthermore, CoVeR encourages the embeddings to be ``topic-aligned'' in that the dimensions have specific independent meanings. This allows our covariate-specific embeddings to be compared by topic, enabling downstream differential analysis. We empirically evaluate the benefits of our algorithm on datasets, and demonstrate how it can be used to address many natural questions about covariate effects.

Accompanying code to this paper can be found at http://github.com/kjtian/CoVeR.
\end{abstract}

\section{Introduction}

The use of factorizations of co-occurrence statistics in learning low-dimensional representations of words is an area that has received a large amount of attention in recent years, perhaps best represented by how widespread algorithms such as GloVe \citep{pennington2014glove} and Word2Vec \citep{mikolov2013distributed} are in downstream applications. In particular, suppose we have a set of words $i \in [n]$, where $n$ is the size of the vocabulary. The aim is to, for a fixed dimensionality $d$, assign a vector $v_i \in \mathbb{R}^d$ to each word in the vocabulary in a way that preserves semantic structure.

In many settings, we have a corpus with additional covariates on individual documents. For example, we might have news articles from both conservative and liberal-leaning publications, and using the same word embedding for all the text can lose interesting information. Furthermore, we suggest that there are meaningful semantic relationships that can be captured by exploiting the differences in these conditional statistics. To this end, we propose the following two key questions that capture the problems that our work addresses, and for each, we give a concrete motivating example of a problem in the semantic inference literature that it encompasses.

\textbf{Question 1:} How can we leverage conditional co-occurrence statistics to capture the effect of a covariate on word usage?

A simple example of a covariate one may wish to condition on is the document the writing came from (i.e. for a group of books, learn a set of word embeddings for each book). There are many natural ways to capture the effect of this conditioning; we choose to do so by representing the covariate as a vector. It is interesting to see if the resulting vectors cluster by author, for example. An example application is addressing the question: did William Shakespeare truly write all the works credited to him, or have there been other ``ghostwriters'' who have contributed to the Shakespeare canon? This is the famous Shakespeare authorship question, for which historians have proposed various candidates as the true authors of particular plays or poems \citep{shakespeare}. If the latter scenario is the case, what in particular distinguishes the writing style of one candidate from another, and how can we infer who the most likely author of a work is from a set of candidates? We show that it is possible to simultaneously address the authorship question, and also learn covariate vectors which are interpretable in terms of their effects on word usage (measured by reweighting the importance of topics). This motivates our next question. 

\textbf{Question 2:} Traditional factorization-based embedding methods are rotationally invariant, so that individual dimensions do not have semantic meaning. How can we break this invariance to yield a model which aligns topics with interpretable dimensions?

There has been much interest in the differences in language and rhetoric that appeal to different demographics. For example, studies have been done regarding ``ideological signatures'' specific to voters by partisan alignment \citep{mindbody} in which linguistic differences were proposed along focal axes, such as the ``mind versus the body'' in texts with more liberal or conservative ideologies. How can we systematically infer topical differences such as these between different communities?

Questions such as these, or more broadly covariate-specific trends in word usage, motivated this study. Concretely, our goal is to provide a general framework through which embeddings of sets of objects with co-occurrence structure, as well as the effects of conditioning on particular covariates, can be learned jointly. As a byproduct, our model also gives natural meaning to the different dimensions of the embeddings, by breaking the rotational symmetry of previous embedding-learning algorithms, such that the resulting vector representations of words and covariates are ``topic-aligned''.

\paragraph{Our Contributions}
Our main contributions are CoVeR, a decomposition algorithm that addresses the goals and issues discussed in the introduction, and the methods for systematic analysis we propose. Namely, we propose a method which pools information between conditional co-occurrence statistics to learn covariate-specific embeddings in a data-efficient way, as well as an interpretable embedding of covariate vectors. We evaluate our method against conditional GloVe as well as the related method of \citep{sefe}.

\paragraph{Paper Organization}
We discuss related works in section 2 and in section 3, we provide our embedding algorithm, as well as mathematical justification for its design. In section 4, we describe our dataset. In section 5, we validate our algorithm with respect to intrinsic properties and standard metrics. In section 6, we propose several experiments for systematic downstream analysis. 

\section{Background and related works}

When designing an algorithm for learning covariate-specific embeddings, we identified three main goals that the algorithm should satisfy. The algorithm should be 1) covariate-specific, 2) data-efficient, and 3) interpretable. Goals 1 and 3 go without saying; goal 2 arises in our setting because oftentimes, the conditional co-occurrence counts can be quite small, especially when there are many covariates. We define all three in very general terms, and use them to evaluate existing methods which can be used to handle our conditional embedding learning task. In this section, we describe several existing approaches and how well they address these goals, which will serve as the basis for our evaluation later in the paper.

\paragraph{Conditional GloVe} 

As a brief introduction to embedding methods, all such algorithms generally rely on the intuition that some function of the co-occurrence statistics is \textit{low rank}.  Studies such as GloVe and Word2Vec proposed  based on minimizing low-rank approximation-error of nonlinear transforms of the co-occurrence statistics. let $A$ be the $n \times n$ matrix with $A_{ij}$ the co-occurrence between words $i$ and $j$, where co-occurrence is defined as the (possibly weighted) number of times the words occur together in a window of fixed length. For example, GloVe aimed to find vectors ${v_i \in \mathbb{R}^d}$ and biases ${b_i \in \mathbb{R}}$ such that the loss
\begin{equation}
J(v, b) = \sum_{i, j = 1}^{n} f(A_{ij})(v_i^T v_j + b_i + b_j - \log A_{ij})^2
\end{equation}
was minimized, where $f$ was some fixed increasing weight function. Word2Vec aimed to learn vector representations via minimizing a neural network-based loss function (it can be shown that Word2Vec and GloVe are essentially performing the same factorization with respect to a different loss function). A related embedding approach is to directly perform principal component analysis on the PMI (pointwise mutual information) matrix of the words \citep{pmifactor}.  PMI-factorization based methods aim to find vectors \{$v_i$\} such that $v_i^T v_j \approx PMI(A)_{ij} = \log \frac{\mathbb{P}(i, j)}{\mathbb{P}(i)\mathbb{P}(j)}$, where the probabilities are taken over the co-occurrence matrix. This is essentially the same  as finding a low-rank matrix $V$ such that $V^T V \approx PMI$, and empirical results show that the resulting embedding captures useful semantic structure.

A simple baseline for learning conditional embeddings thus is simply performing GloVe (or another non-conditional embedding learning method) on each conditional co-occurrence matrix, and using the resulting vectors as our conditional embedding. This certainly satisfies goal 1 that we defined earlier, but neither goal 2 nor goal 3 is addressed. Regarding goal 2, this naive method does not pool information between the co-occurrence matrices, and thus is inefficient in its learning of semantic meaning, which is very relevant when the conditional counts are not large (for example, if the covariate is time intervals, the conditional counts can be very small), which we show empirically. Regarding goal 3, because of the rotational invariance of the method, there is no straightforward way to relate the conditional embeddings to one another. There has been previous work which learns a different set of embeddings on each conditional slice and then tries to postprocess to align these embeddings, such as \citep{temporal}. However, these works still run into the data efficiency issues we describe, and it is difficult to analyze the quality of the alignment (for example, the embedding algorithm on different slices of the tensor may converge to minima which are not related by a rotation, therefore an alignment may not exist).


\paragraph{Tensor-based Conditional Embedding Methods}

An interesting recent work related to our method is the conditional embedding method S-EFE \citep{sefe}. Their work also aims to learn a different set of word embeddings for each covariate (the same problem we consider), and does so via a neural-net based objective function. Their work is analogous to our work up to the parameterization of the covariates (our work parameterizes covariates via a diagonal scaling matrix, and theirs by a small neural network). This is similar in some ways to how Word2Vec and GloVe are analogous, but very different in others: the primary difference between Word2Vec and GloVe is in the choice of loss between distributions in the objective, and our work differs from S-EFE more fundamentally in the parameterization. Furthermore, this work does a sub-optimal job of addressing goals 2 (due to the large number of parameters necessary in a neural-net based objective) and 3 (interpreting parameters in a nonlinear transform is difficult), which we show in evaluations.

There are a few other related works which also try to learn grouped embeddings via tensor decomposition \citep{tensor1} \citep{tensor2}. However, these works consider conditioning in a restricted sense (typically with respect to conditioning on different semantic meanings of a word), do not aim to interpret the context vectors, and do not try to preserve rotational alignment with interpretable dimensions. In particular, because these models do not consider the direct representation of the covariate in the objective function, the representations are missing key structural properties (for example topic alignment), which doesn't allow for meaningful downstream analysis. In this sense, our work aims to solve a more general problem, and provides further justification for tensor-based embedding methods.

\paragraph{Multi-sense Embeddings}

There has been work which aims to address the multiple-meaning problem which arises in learning word embeddings, via multi-sense embeddings \citep{multisense1} \citep{multisense2}. Typically, these algorithms work by allocating multiple vectors for each word (either a fixed count or a varying number), and optimizing an objective function which picks out one of the vectors for each word. For example, if words $i$ and $j$ occur in the same context, the corresponding term in the objective function might use the embeddings (or senses) of $i$ and $j$ which are closest to each other. In some ways, this is similar to our idea that the same word can have different meanings in different contexts. However, it is addressing a fundamentally different problem. The ``senses'' which are learned correspond to differing meanings of a word, whereas our work relies on the intuition that across contexts, a word will have roughly similar meanings, but usage may differ in some ways, perhaps with respect to certain topics. Furthermore, while multi-sense embeddings pose an interesting preliminary way to address our problem, they are certainly neither covariate-specific (goal 1), nor interpretable (goal 3, in that the different learned sense vectors do not need to be related to each other for the same word).

\section{CoVeR: Motivation and Embedding Algorithm}

\paragraph{Notation}
Throughout this section, we will assume a vocabulary of size $n$ and a discrete covariate to condition on, where the covariate can take on $m$ values (for example, if the covariate is the community that the corpus comes from, i.e. liberal or conservative discussion forums, $m$ is simply the number of communities). It is easy to see how our algorithm generalizes to higher-order tensor decompositions when there are multiple dimensions covariates to condition on (for example, slicing along community and slicing along timeframe simultaneously). Words will be denoted with indices $i, j \in [n]$ and covariates with index $k \in [m]$. All embedding vectors will be in $\mathbb{R}^d$, and dimensions in our embedding are referred to by index $t \in [d]$.t

We will denote the co-occurrence tensor as $A \in \mathbb{R}^{n \times n \times m}$, where $A_{ijk}$ denotes how many times words $i$ and $j$ occurred together within a window of some fixed length, in the corpus coming from covariate $k$. The result of our algorithm will be two sets of vectors, \{$v_i \in \mathbb{R}^d$\} and \{$c_k \in \mathbb{R}^d$\}, as well as bias terms that also fit into the objective. Finally, let $\odot$ denote the element-wise product between two vectors.

\paragraph{Objective Function and Discussion}
 Here, we give the objective function our method minimizes, and provide some explanation for how one should imagine the effect of the covariate weights. 
The objective function we minimize is the following \textit{partial non-negative tensor factorization} objective function for jointly training word vectors and weight vectors representing the effect of covariates, adapted from the original GloVe objective (note that $c_k \odot v_i = \textbf{diag}(c_k) v_i$, where $\textbf{diag}(c_k)$ is the diagonal matrix weighting of covariate $k$), which we call $J(v, c, b)$:
\begin{equation}
\sum_{i, j = 1}^{n} \sum_{k = 1}^{m} f(A_{ijk})((c_k \odot v_i)^T (c_k \odot v_j) + b_{ik} + b_{jk} - \log A_{ijk})^2
\end{equation}
which is to be optimized over \{$v_i \in \mathbb{R}^d$\}, \{$c_k \in \mathbb{R}^d$\}, and \{$b_{ik} \in \mathbb{R}$\}. To gain a little more intuition for why this is a reasonable objective function, note that the resulting objective for a single ``covariate slice'', $J_k(v, c, b)$, is essentially
\begin{equation}
\sum_{i, j = 1}^{n} f(A_{ijk})((c_k \odot v_i)^T (c_k \odot v_j) + b_{ik} + b_{jk} - \log A_{ijk})^2
\end{equation}
which fits the vectors $c_k \odot v_i$ to the data, thus approximating the statistic $\log A_{ijk}$ with $\sum_{t = 1}^d v_{it} v_{jt} c_{kt}^2$. Note that in the case $m = 1$, the model we use is identical to the standard GloVe model since the $c_k$ can be absorbed into the $v_i$. We used $f(x) = (\frac{\min(100, x)}{100})^{0.75}$, to parallel the original objective function in \citep{pennington2014glove}.

One can think of the dimensions our model learns as independent topics, and the effects of the covariate weights $c_k$ as up- or down-weighting the importance of these topics in contributing to the conditional co-occurrence statistics.

\paragraph{A Geometric View of Embeddings and Tensor Decomposition}
We provide a geometric perspective on our model. 
Throughout this section, note at a high level, the aim of our method is to learn sets \{$v_i \in \mathbb{R}^d$\}, \{$c_k \in \mathbb{R}^d$\}, and \{$b_{ik} \in \mathbb{R}$\} for $1 \leq i \leq n, 1 \leq k \leq m$, such that for a fixed $k$, the vectors \{$c_k \odot v_i$\} and the biases \{$b_{ik}$\} approximate the vectors and biases learned from running GloVe on only the $k^{th}$ slice of the co-occurrence tensor.

We now provide some rationale for why this is a reasonable objective. The main motivation for our algorithm is that under standard distributional assumptions, the uniform distribution that word vectors are sampled from can be seen as a sphere \cite{arora2015rand}, which has been experimentally verified. A natural way to model the effect of conditioning on a covariate is to replace this spherical distribution with an ellipse, which is the equivalent of reweighting the dimensions of the word vectors with respect to some basis. We model this effect as multiplying the embedding vectors themselves by a symmetric PSD matrix, specific to the covariate which is being conditioned on. In this framework, assign each covariate $k$ its own symmetric PSD matrix, $B_k$. It is well-known that any symmetric PSD matrix can be factorized as $B_k = R_k^T D_k R_k$ for some orthonormal basis $R_k$ and some (nonnegative) diagonal $D_k$; it thus suffices to consider the effect of a covariate on some ground truth ``base embedding'' $M$ as applying the linear operator $B_k$ to each embedding vector, resulting in the new embedding $B_k M$.


\begin{figure}[ht!]
\begin{center}
\includegraphics[width=.32\textwidth]{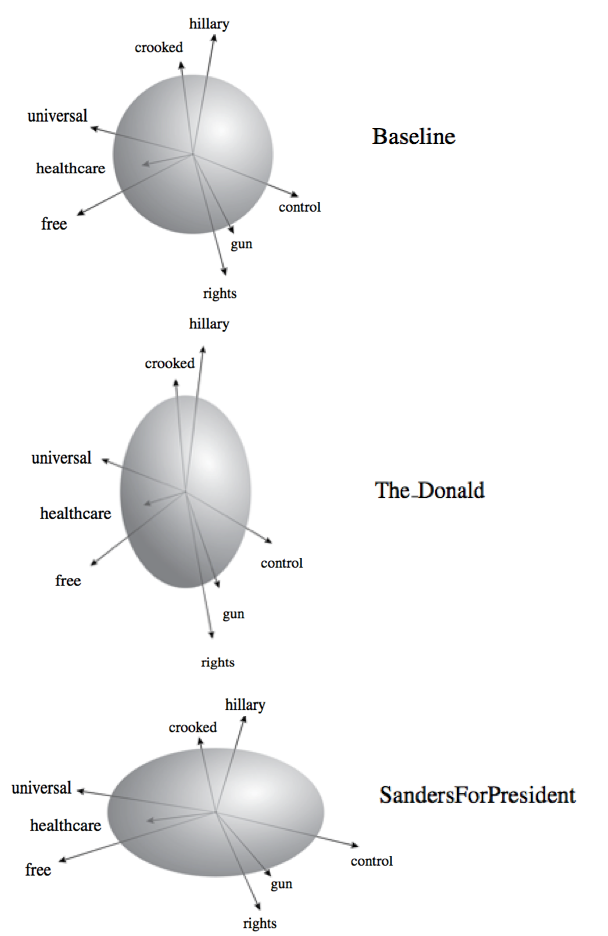}
\end{center}
 \caption{The effects of conditioning on covariates (covariates are discussion forums, described in Section 4). Left: baseline embedding with some possible word embedding positionings. Middle, right: embedding under effect of covariates. For example, ``hillary'' and ``crooked'' are pushed closer together under effects of \textit{The\_Donald}, and ``healthcare'' pushed closer to ``universal'' and ``free'' under effects of \textit{SandersForPresident}. ``Gun'' is moved closer to ``rights'' and further from ``control'' under \textit{The\_Donald}, and vice versa in \textit{SandersForPresident}.\label{fig:spheres}}
 
\end{figure}

This model is quite expressive in its own right, but we consider a natural restriction where there exists one basis $R$ under which the resulting embeddings are affected by covariates via multiplication by a diagonal matrix, instead of a PSD matrix (Fig.\ref{fig:spheres}). In particular, we note that 
\begin{equation}
M^T B_k^T B_k M = M^T R_k^T D_k R_k R_k^T D_k R_k M = M_k'^T D_k^2 M_k'
\end{equation}
where $M_k' = R_k M$ is a rotated version of $M$. Now, in the restricted model where all the $R_k$ are equal, we can write all the $M_k'$ as $RM$, so it suffices to just consider the rotation of the basis that the original embedding was trained in where $R$ is just the identity (since matrix-factorization based word embedding models are rotation invariant). Under this model, the co-occurrence statistics under some transformation should be equivalent to $M^T D_k^2 M$. Now, clearly this is the same as finding a scaling vector $c_k$ (such that $D_k$ is \textbf{diag}$(c_i)$) and using the vectors $v_i \cdot c_k$ in the factorization objective function, exactly which is implied by equation 4.

Note that this is essentially saying that in this distributed word representation model, there exists some rotation of the embedding space under which the effect of the covariate separates along dimensions. The implication is that there are some set of independent ``topics'' that each covariate will upweight or downweight in importance (or possibly ignore altogether with a weight of 0), characterizing the effect of the conditioning directly in terms of the effect on topics.

\paragraph{Algorithm Details}
Our model learns the resulting parameters \{$v_i \in \mathbb{R}^d$\}, \{$c_k \in \mathbb{R}^d$\}, and \{$b_{ik}$\}, by using the Adam \citep{kingma2014adam} algorithm, which was empirically shown to yield good convergence results in the original GloVe setting. The specific hyperparameters used for each dataset will be described in the next section.
All vectors were initialized as random unit vectors \footnote{We also experimented with initializing covariate weight vectors as random vectors centered around the all 1 vector. This initialization also yielded the sparsity patterns discussed in the next section, but converged at a slower rate, and performed similarly on downstream metrics as initializing near all 0, so we kept this initialization.}.

\section{Dataset}
We evaluated CoVeR in two primary datasets. 
Co-occurrence statistics were formed by considering size 8 windows and using an inverse-distance weighting (e.g. words 3 apart had $\frac{1}{3}$ added), which was suggested by some implementations of \citep{pennington2014glove}.

The first dataset, referred to as the ``book dataset'', consists of the full text from 29 books written by 4 different authors. The books we used were J.K. Rowling's ``Harry Potter'' series (7 books), ``Cormoran Strike'' series (3 books), and ``The Casual Vacancy''; C. S. Lewis's ``The Chronicles of Narnia'' series (7 books), and ``The Screwtape Letters''; George R. R. Martin's ``A Song of Ice and Fire'' series (5 books); and Stephenie Meyer's ``Twilight'' series (4 books), and ``The Host''. These books are fiction works in similar genres, with highly overlapping vocabularies and common themes. A trivial way of learning series-specific tendencies in word usage would be to cluster according to unique vocabularies (for example, only the ``Harry Potter'' series would have words such as ``Harry'' and ``Ron'' frequently), so the co-occurrence tensor was formed by looking at all words that occurred in all of the series with multiple books, which eliminated all series-specific words. Furthermore, series by the same author had very different themes, so there is no reason intrinsic to the vocabulary to believe the weight vectors would cluster by author. The vocabulary size was 5,020, and after tuning our algorithm to embed this dataset, we used 100 dimensions and a learning rate of $10^{-5}$.

The second dataset, the ``politics dataset'', was a collection of comments made in 2016 in 6 different subreddits on the popular discussion forum reddit, and was selected to address both Questions 1 and 2. The covariate was the discussion forum, and the subreddits we used were \textit{AskReddit}, \textit{news}, \textit{politics}, \textit{SandersForPresident}, \textit{The\_Donald}, and \textit{WorldNews}. \textit{AskReddit} was a baseline discussion forum with a very general vocabulary usage, and the discussion forums for the Sanders and Trump support bases were also selected, as well as three politically-relevant but more neutral communities (it should be noted that the \textit{politics} discussion forum tends to be very left-leaning). We considered a vocabulary of size 15,000, after removing the 28 most common words (suggested by personal communication amongst the word embedding community) and entries of the cooccurrence tensor with less than 10 occurrences (for the sake of training efficiency). The embedding used 200 dimensions and a learning rate of $10^{-5}$.

\section{Experimental Validation}
\subsection{Clustering by Weights}

We performed CoVeR on the book dataset, and considered how well the weight vectors of the covariate clustered by series and also by author. We also considered clustering of the covariate parameter vectors learned from the S-EFE method. As a baseline, we considered the topic breakdown on the same co-occurrence statistics predicted by Latent Dirichlet Allocation \citep{lda} with various dimensionalities (20, 50, 100). For a visualization, see Fig. 2.

\begin{figure}[ht!]
\centering
\begin{subfigure}{.3\textwidth}
	\centering
    \includegraphics[width=\linewidth]{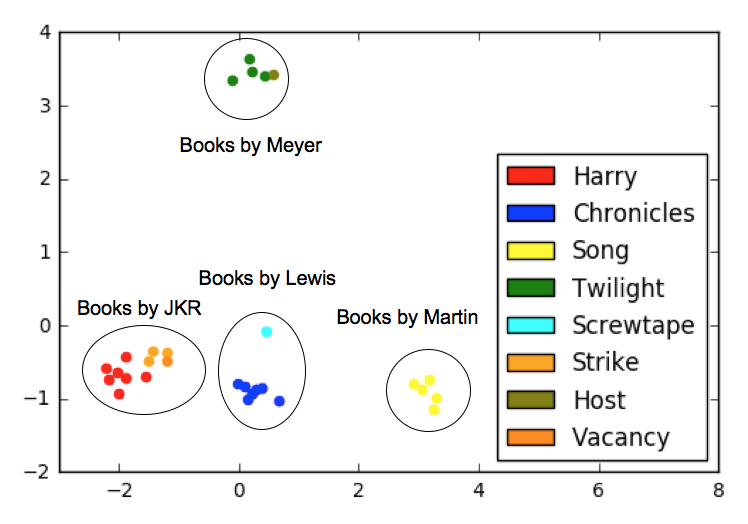}
    \caption{2D PCA of book dataset weight vectors (CoVeR, 100 dimensions)}
    \label{fig:sub1}
\end{subfigure}
\begin{subfigure}{.3\textwidth}
	\centering
    \includegraphics[width=\linewidth]{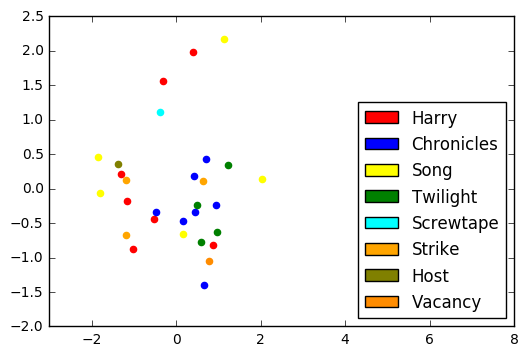}
    \caption{2D PCA of neural net parameter vectors (S-EFE, 100 dimensions).}
    \label{fig:sub1}
\end{subfigure}
\begin{subfigure}{.3\textwidth}
	\centering
    \includegraphics[width=\linewidth]{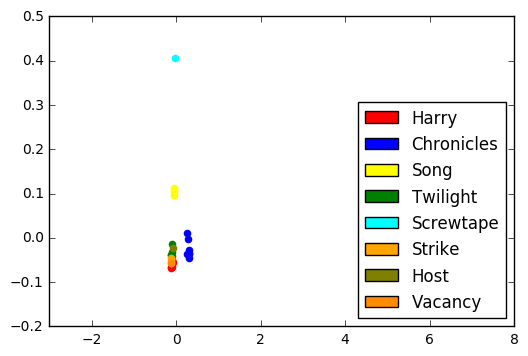}
    \caption{2D PCA of book dataset topic vectors, as predicted by LDA (100 topics)}
    \label{fig:sub2}
\end{subfigure}
\caption{Visualization of topic vectors, algorithm comparison}
\label{fig:sparsity}
\end{figure}

For every book in every series, the closest weight vectors (via the tensor decomposition algorithm) by series were all the books in the same series. Furthermore, for every book written by every author, the closest weight vectors by author were all the books by the same author. This clear clustering behavior even when only conditioning on co-occurrence statistics for words that appear in all series (throwing out series-specific terms) implies the ability of the weight vectors to cluster according to higher-order information, perhaps such as writing style. 

In contrast, S-EFE is does not learn interpretable features for each covariate; simply plotting the S-EFE parameter vector does not cluster the books by the same author. We recognize that the comparison of the PCA-projected S-EFE parameter vectors should not be expected to do well (we tried nonlinear dimension reduction techniques such as T-SNE under various parameter settings as well, and in no setting did the projections cluster meaningfully); nonetheless, using the parameter vectors seemed to be the most natural way of measuring interpretability with respect to the covariates. All parameter settings of LDA failed to produce any meaningful clusters.

\subsection{Specificity of Conditional Embeddings}

When measuring comparability across embeddings, one desirable feature of an embedding algorithm is that it redistances words with respect to how much the meaning changes across covariates. For example, a common word such as ``the'' or ``a'' has essentially the same meaning in almost every situation, so the resulting embedding should not change very much. However, for a word that may have very different meanings in different contexts (such as ``immigration'', which might have different connotations in liberal vs. conservative texts), we expect the covariate-specific embeddings to be more spread out.

A simple way to quantitatively evaluate this notion of ``variance in meaning = spread in embeddings'' is to, for each word, consider the average pairwise cosine distance between covariate-specific embeddings. For example, for the word ``dark'', using the book dataset, there will be 29 different embeddings corresponding to each different set of covariate cooccurrences; over all pairs of books, take the average cosine distance between the 
embeddings of the given word. We plotted the distribution of this average across all words.

\begin{figure}[ht]
  \label{table:cosine-similarity}
  \centering
  \includegraphics[width=.75\linewidth]{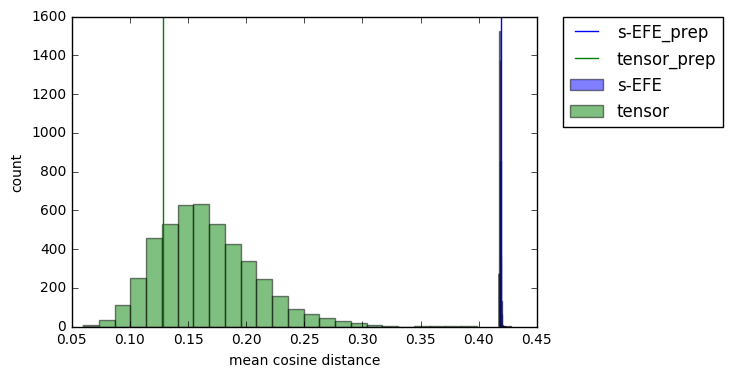}
    \caption{Histogram of average pairwise cosine similarity for context-specific embeddings of a given word.}
\end{figure}

The results are quite striking: neural embedding method such as S-EFE is unable to distinguish between the specificities of words. This could be a result of the neural net parameterization learning a different subspace for each covariate, which makes comparing embeddings across groups infeasible. However, CoVeR is able to relate the covariate-specific embeddings to each other via a linear re-weighting, giving meaning to comparisons between the embeddings. As a baseline, the mean distance for common prepositions (which we take to represent ``stable'' words which should not drift too much under conditioning) is plotted; they were much more stable in our embedding, as expected.

\subsection{Data Efficiency and Validation}

Consider the problem of learning an embedding for the text of a particular book (or series). This is fundamentally a covariate-specific embedding problem, so we test the data efficiency of our algorithm with respect to two alternative methods of learning conditional embeddings. The main advantage given by using CoVeR over applying GloVe to the individual slices is one of data efficiency: by pooling the co-occurrence statistics of words across other books, we are able to give less noisy estimates of the vectors, especially for rare or nonexistent words. We also show that our method compares very favorably with respect to S-EFE. Individual books contained between 26747 and 355814 words.

To this end, we performed the following experiment. For some book $k$, consider three embeddings: 1) the result of performing GloVe on the co-occurrence statistics of just the book, 2) the context-specific embedding produced by S-EFE with dimension 100 (different dimensionalities compared similarly) 3) the (weighted) context-specific embedding resulting from CoVeR. Then, we tested these resulting embeddings using a standard suite of evaluation metrics \citep{evaluation}, including cluster purity and correlation similarities. Our method outperformed method 1 on all tasks and method 2 on all but one, often significantly.

\begin{table}[ht]
\small
  \caption{Average performance on Harry Potter books, cluster purity. Larger scores indicate better performance.}
  \label{table:standard-test}
  \centering
  \begin{tabular}{c|ccc}
    \toprule
     \multicolumn{1}{c|}{ }  &
    \multicolumn{3}{c|}{Cluster purity}\\
    &	AP	& BLESS & Battig \\
    \midrule
  CoVeR	&\textbf{0.1602} & \textbf{0.2185} & \textbf{0.0877} \\
  S-EFE & 0.1289 & 0.1942 & 0.0745 \\
GloVe &0.1297 &	0.2042 & 0.0770 \\
    \bottomrule
  \end{tabular}
\end{table}

\begin{table}[ht]
\small
  \caption{Average performance on Harry Potter books, correlation similarities. Larger scores indicate better performance.}
  \label{table:standard-test}
  \centering
  \begin{tabular}{c|cccc}
    \toprule
     \multicolumn{1}{c|}{ }  &
    \multicolumn{4}{c|}{Correlation similarities} \\
    &	MEN	& MTurk & RG65	& RW \\
    \midrule
  CoVeR	& \textbf{0.1370} & \textbf{0.0719} & \textbf{0.1272} & \textbf{0.0902} \\
  S-EFE & 0.0184 & 0.0986 & -0.0098 & -0.0753 \\
GloVe & 0.0593 & 0.0341 & 0.0133 & 0.0588 \\
    \bottomrule
  \end{tabular}
\end{table}

\subsection{Sparsity of Weight Vectors}

Because of experimentally verified isotropic distributional properties of word embedding vectors \citep{arora2015rand}, it is unreasonable to ask for sparsity in the word embeddings themselves. However, our topic-specific weighting scheme for the covariate weight vectors implies that sparsity is desirable in the weights. The weight sparsity resulting from our algorithm was experimentally verified through many runs of our embedding method, as well as across different optimization methods, which to us was a surprising and noteworthy artifact of our algorithm.
\begin{figure}[h]
\centering
    \includegraphics[width=.8\linewidth]{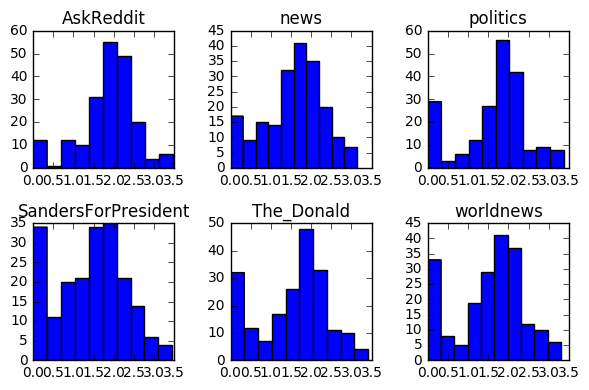}
    \caption{Histogram, sizes of weights in covariate vectors for politics dataset. Smallest bucket is zero ($< 10^{-10}$). Sparsity of covariates in book dataset deferred to appendix.}
\label{fig:sparsity}
\end{figure}

Note that in the objective function (3), sparse coordinates will become ``stuck'' at 0, because the gradient update $\frac{\partial J}{\partial c_{kt}}$ of $c_{kt}$ is proportional to $c_{kt}$:
\begin{equation}
\sum_{i, j = 1}^{n} 4f(A_{ijk})((c_k \odot v_i)^T (c_k \odot v_j) + b_{ik} + b_{jk} - \log A_{ijk})c_{kt}
\end{equation}
The dimensions that the covariates were sparse in did not overlap by much: on average, weight vectors had 20.7 sparse coordinates, and the average number of coordinates that was sparse in both vectors of a random pair was 5.2. This suggests that the set of ``topics'' each conditional slice of the tensor does not care about is fairly specific. Experimenting with regularizing the word vectors forced the model into a smaller subspace (i.e. sparse dimensions existed but were shared across all the words), which is not useful and provides further evidence for a natural isotropic distribution.

\begin{table}[ht]
\small
  \caption{Sparse coordinate count, covariate weight vectors, 5 runs.}
  \label{table:sparsity-patterns}
  \centering
  \begin{tabular}{c|c|c|c|c|c|c}
  & \textit{Ask} & \textit{news} & \textit{politics} & \textit{Sanders} & \textit{Donald} & \textit{world}\\
mean &  18.2 &  16.8 &  25.6 &  36.8 & 27.2 &  23.2 \\
std & 1.4 &  2.4 & 2.2&  3.6 &2.6 &3.1 \\
  \end{tabular}
\end{table}

To confirm that the sparsity was a result of separation of covariate effects rather than an artifact of our algorithm, we ran our decomposition on a co-occurrence tensor which was the result of taking the same slice (subreddit) and subsampling its entries 3 times, creating 3 slices of essentially similar co-occurrence statistics. We applied our algorithm with different learning parameters, and the resulting weight vectors after 90 iterations (when the outputs vectors converged) were extremely non-sparse, with between 0 and 2 sparse coordinates per weight vector. The dimensions that are specifically 0 for a covariate corresponds to topics that are relatively less relevant for that covariate. In the next section, we develop methods to systematically interpret the covariate weights in terms of topics.

\section{Interpretation}

\subsection{Inference of Topic Meaning}

One interesting problem to the word embedding community is that of topic inference and modeling, that is discovering underlying topics via analyzing co-occurrence statistics over corpora \citep{topicmodel}. A simple test of inferring topic meaning (i.e. topics coordinates are associated with) is to consider the set of words which are large in the given coordinate. Concretely the task is, given some index $t \in [d]$, output the words whose (normalized) vectors have largest absolute value in dimension $t$. We show the results of this experiment for several of the sparse coordinates in the \textit{AskReddit} weight vector:

\begin{table}[ht]
\begin{center}
  \caption{Representative sample of top words for selected dimensions: topic inference task. Each word was in top 20 normalized coordinate values for the dimension $t$.}
  \small
  \begin{tabular}{|l|l|l|}
  \hline
  $t$ & Highest weighted words, dimension $t$ & Meaning \\
  \hline
  99 & horses, cat, teenager, grandma & people/animals\\
  \hline
  120 & tables, driveway, customer, stations & domestic \\
  \hline
  183 & gate, territory, directions, phillipines & foreign \\
  \hline
  194 & sweat, disciplined, beliefs, marines & military \\
  \hline
  \end{tabular}
\end{center}
\end{table}

There are several conclusions to be drawn from this experiment. Firstly, while there is some clear noise, specific topic meanings do seem to appear in certain coordinates (which we infer in the table above). It is reasonable that meaning would appear out of coordinates which are sparsely weighted in some covariate, because this means that it is a topic that is relevant in other discussion forums but purposely ignored in some specific forum, so it is likely to have a consistent theme. When we performed this task for coordinates which had low variance in their weights between covariates, the resulting words were much less consistent.

It is also interesting to see how covariates weight a topic with an identified meaning. For example, for coordinate 194 (military themes), \textit{AskReddit} placed negligible weight, \textit{news} and \textit{worldnews} placed weight 2.06 and 2.04, \textit{SandersForPresident} and \textit{The\_Donald} placed weight 0.41 and 0.38, and \textit{politics} placed weight 0.05. This process can also be reversed - for example, consider coordinates small in \textit{worldnews} and large in \textit{news}. One example was coordinate 188; upon checking the words large in this coordinate (\{taser, troops, supremacists, underpaid, rioters, amendment, racially\}) it clearly had themes of rights and protests, which makes sense as a domestic issue, not a global one. 

\subsection{Topical Word Drift}

We performed the following experiment: which pairs of words start off close in the baseline embedding, yet under some covariate weights move far apart (or vice versa)? Concretely, the task is, for a fixed word $i$ and a fixed covariate $k$, to identify words $j$ such that $||c_k \odot v_i - c_k \odot v_j|| \gg ||v_i - v_j||$ or $||c_k \odot v_i - c_k \odot v_j|| \ll ||v_i - v_j||$, where the magnitude of drift is quantified by the ratio of distances in the normalized embedding. The motivation is to find words whose general usage is similar, but have very different meanings in specific communities. We present a few representative examples of this experiment, for $k$ = \textit{The\_Donald}.

\begin{table}[ht]
\begin{center}
  \small
  \caption{Representative examples of word drift for \textit{The\_Donald}.}
  \begin{tabular}{|l|l|l|}
  \hline
  Word $i$ & Drift dir. & Words with strongest drift \\
  \hline
  hillary & Closer & crooked, lying, shillary, killary \\
  \hline
  hillary & Further & electable, compromise, favored \\
  \hline
  gun & Closer & merchandise, milo, flair, fakenews \\
  \hline
  gun & Further & assault, child, fanatics, problem, police \\
  \hline
  immigrant & Closer & unauthorized, uninsured, parasite \\
  \hline
  immigrant & Further & child, abused, future, attorneys, protect \\
  \hline
  \end{tabular}
\end{center}
\end{table}


Combining the previous two sections allows us to do an end-to-end case study on words that drift under a covariate, so we can explain specifically which topics (under reweighting) caused this shift. For example, the words ``immigrant'' and ``parasite'' were significantly closer under the weights placed by \textit{The\_Donald}, so we considered dimensions that were simultaneously large in the vector $v_{immigrant} - v_{parasite}$ and sparse in the weight vector $c_{The\_Donald}$. The dimensions 89 and 139 were sparse and also the 2nd and 3rd largest coordinates in the difference vector, so they had a large contribution to the subspace which was zeroed out under the reweighting. Words that were large in these dimensions (and thus representative of the zeroed out subspace meaning) included \{misunderstanding, populace, scapegoat, rebuilding\} for 89, and \{answers, barriers, applicants, backstory, selfless, indigenous\} for 139. This suggests two independent reasons for the drift: dimensions corresponding to emotional appeal and legal immigration being zeroed out.

\subsection{Covariate-Specific Analogies}

One of the most famous downstream applications of recent embedding methods such as \citep{pennington2014glove} and \citep{mikolov2013distributed} is representing analogies. This is formulated as $a$ is to $b$ as $c$ is to $d$ $\leftrightarrow v_a - v_b \approx v_c - v_d$, for example $v_{woman} - v_{queen} \approx v_{man} - v_{king}$. We considered how well our method captured \textit{covariate-specific analogies}, which appear in a covariate-specific embedding but not most others. Our embedding was able to capture differential meaning in the form of these specific analogies: we describe the results more formally in the appendix. An example is that 
``hillary is to liberal as trump is to (white / racist)'' are both much stronger analogies in liberal discussion forums, and a corresponding strong analogy in conservative forums is ``... as trump is to disenfranchised''. Also, across the board, ``hillary is to politician as trump is to businessman'', but replacing ``businessman'' with ``confidence'' is an enriched analogy in conservative forums, and replacing with ``irrationality'' is enriched in liberal ones.

\paragraph{Discussion} We have presented a joint tensor model that learns an embedding for each word and for each covariate. This makes it very simple to compute the covariate specific embedding: we just take the element-wise vector product. It also enables us to systematically interpret the covariate vector by looking at dimensions along which weight is large or 0. Our experiments show that these dimensions can be interpreted as coherent topics. While we focus on word embeddings, our tensor covariate embedding model can be naturally applied in other settings. For example, there is a large amount of interest in learning embeddings of individual genes to capture biological interactions. The natural covariates here are the cell types and our method would be able to model cell-type specific gene interactions. Another interesting setting with conditional covariates would be time-series specific embeddings, where data efficiency becomes more of an issue. We hope our framework is general enough that it will be of use to practitioners in these settings and others. 
\newpage
\section*{Acknowledgements}

This research was supported by NSF Graduate Fellowship DGE-1656518. The authors thank Tatsunori Hashimoto for helpful comments and Lilly Shen for helping with illustrations.

\newpage

\bibliography{example_paper}
\bibliographystyle{icml2018}

\newpage

\onecolumn
\appendix
\onecolumn
\section{Covariate-specific analogies}
We wish to learn covariate-specific analogies of the form $a$ is to $b$ as $c$ is to $d$. To this end, we considered experiments of the form: for fixed words $a, b, c$, determine words $d$ such that for some covariate $k$, the quantity

\begin{equation}
\frac{(c_k \odot v_a - c_k \odot v_b) \cdot (c_k \odot v_c - c_k \odot v_d)}{||c_k \odot v_a - c_k \odot v_b|| ||c_k \odot v_c - c_k \odot v_d||}
\end{equation}

is small, yet for other $k$, the quantity is large. The intuition is that under the covariate transform, $v_c - v_d$ points roughly in the same direction as $v_a - v_b$, and $d$ is close to $c$ in semantic meaning.

In particular, we set $a$ = ``hillary'', $c$ = ``trump'', and found words $b$ for which there existed a $d$ consistently at the top across subreddits (implying existence of strong analogies). For example, when $b$ = ``woman'', $d$ = ``man'' was the best analogy for every weighting. Then, for these $b$, we considered words $d$ whose relative rankings in the subreddits had high variance. The differential analogies captured were quite striking: the experiment is able to reveal words whose relative meaning in relation to anchor words such as ``hillary'' and ``trump'' drifts significantly.

\begin{table}[ht]
\begin{center}
  \caption{Analogies task. Each best analogy $d$ was one of the top-ranked words in every embedding. We present words whose ``relative analogy'' rank was enriched in some embedding. Subreddits are color-coded: green for news-related WN and N (\textit{worldnews}, \textit{news}), blue for left-leaning P and S (\textit{politics}, \textit{SandersForPresident}), red for right-leaning D (\textit{The\_Donald}), black for A (\textit{AskReddit}).}

\begin{tabular}{|l|l|l|l|l|}
\hline
Word $b$                      & Best analogy $d$                & Word            & High rank         & Low rank              \\ \hline
\multirow{6}{*}{woman}      & \multirow{6}{*}{man}          & abysmal         & 1351 (\textcolor{blue}{S}), 2218 (\textcolor{blue}{P})  & 14329 (base), 14077 (\textcolor{red}{D}) \\ \cline{3-5} 
                            &                               & amateur         & 1543 (\textcolor{blue}{P}), 3966 (\textcolor{blue}{S})  & 13840 (base), 13734 (\textcolor{red}{D}) \\ \cline{3-5} 
                            &                               & zionist         & 1968 (\textcolor{blue}{P}), 2327 (\textcolor{blue}{S})  & 14173 (base), 14248 (A) \\ \cline{3-5} 
                            &                               & politician      & 2796 (\textcolor{green}{WN}), 3155 (\textcolor{red}{D}) & 11959 (base), 10386 (\textcolor{blue}{S}) \\ \cline{3-5} 
                            &                               & president       & 2452 (\textcolor{red}{D}), 3564 (\textcolor{green}{WN}) & 12257 (base)           \\ \cline{3-5} 
                            &                               & nationalists    & 208 (\textcolor{blue}{S}), 606 (\textcolor{red}{D})    & 8916 (base), 7526 (A)   \\ \hline
\multirow{2}{*}{democrat}   & \multirow{2}{*}{republican}   & south           & 3511 (\textcolor{blue}{P})           & 11091 (base)           \\ \cline{3-5} 
                            &                               & bigot           & 400 (\textcolor{blue}{S}), 530 (A)    & 12888 (\textcolor{red}{D}), 12994 (\textcolor{green}{WN})   \\ \hline
\multirow{5}{*}{liberal}    & \multirow{5}{*}{conservative} & christian       & 33 (\textcolor{blue}{P})             & 12756 (\textcolor{red}{D}), 12722 (\textcolor{green}{WN})   \\ \cline{3-5} 
                            &                               & white           & 619 (\textcolor{blue}{P})            & 12824 (\textcolor{red}{D}), 13273 (base) \\ \cline{3-5} 
                            &                               & racist          & 252 (\textcolor{blue}{P})            & 12930 (\textcolor{blue}{S}), 12779 (\textcolor{red}{D})    \\ \cline{3-5} 
                            &                               & sociopathic     & 1756 (\textcolor{red}{D})           & 13744 (\textcolor{blue}{P}), 13389 (A)    \\ \cline{3-5} 
                            &                               & disenfranchised & 3693 (\textcolor{red}{D})           & 11267 (base)           \\ \hline
\multirow{3}{*}{politician} & \multirow{3}{*}{businessman}  & confidence      & 2768 (\textcolor{red}{D})           & 10528 (base)           \\ \cline{3-5} 
                            &                               & questionable    & 598 (\textcolor{green}{WN})           & 13002 (base)           \\ \cline{3-5} 
                            &                               & irrational      & 2153 (\textcolor{green}{N}), 3430 (\textcolor{blue}{P})  & 13305 (base)           \\ \hline
\end{tabular}
\end{center}
\end{table}

\section{Sparsity results: book dataset}

\begin{figure}[h]
\centering
    \includegraphics[width=1\linewidth]{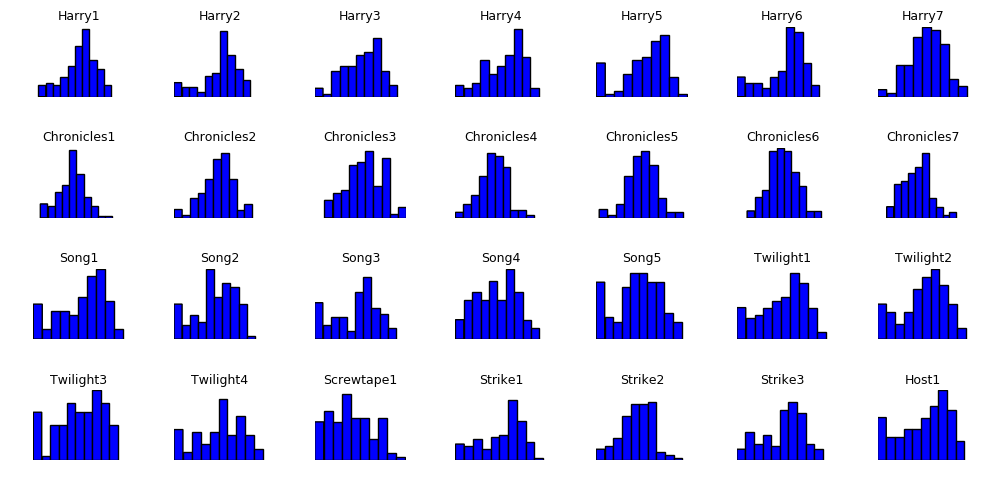}
    \caption{Histogram, sizes of weights in covariate vectors of book data.}
\label{fig:sparsity_book}
\end{figure}

Number of sparse coordinates (out of 100) were as follows, by series and then book order: Harry (0, 5, 1, 3, 7, 4, 0), Chronicles (0, 0, 0, 1, 0, 0, 0), Song (8, 8, 9, 4, 11), Twilight (6, 6, 8, 7), Screwtape (5), Strike (3, 2, 2), Host (6), Vacancy (4). While not as dramatic as in the politics dataset, the presence of zero (rather than small) coordinates across multiple runs shows that there still is specificity of topics being learned. We plot the histogram of coordinate sizes in the following figure.

\section{Algorithm setting notes}

\begin{figure}[ht!]
\centering
\begin{subfigure}{.47\textwidth}
	\centering
   \includegraphics[width=\linewidth]{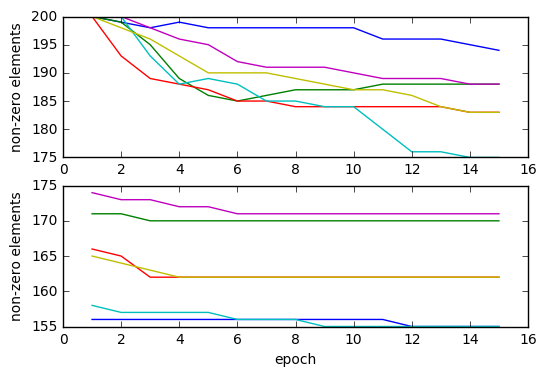}
   \caption{Non-zero dimensions in weight vectors by epoch and optimization method. Upper: Adam; lower: Adagrad.}
\end{subfigure}
\begin{subfigure}{.47\textwidth}
	\centering
    \includegraphics[width=\linewidth]{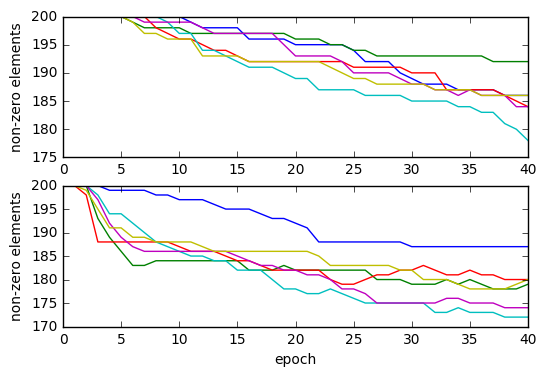}
    \caption{Non-zero dimensions in weight vectors by epoch. Upper: initialization centered around all-1 vector; lower: centered around all-0 vector.}
\end{subfigure}
\label{fig:alg-setting}
\end{figure}

We also experimented with using \citep{adagrad} as the optimization method, but the resulting weight vectors in the politics dataset had highly-overlapping sparse dimensions. This implies that the optimization method tried to fit the model to a smaller-dimensional subspace, which is not a desirable source of sparsity.


%



\end{document}